\documentclass[10pt,conference]{IEEEtranmod}
\usepackage[usenames,dvipsnames]{color}
\usepackage{graphicx}
\usepackage{amsmath,amssymb} 
\usepackage{amssymb}
\usepackage[colorlinks=true,urlcolor=blue,linkcolor=black,citecolor=blue,bookmarks=true,pdfauthor={ZongYuan Ge, Chris McCool, Conrad Sanderson, Peng Wang, Lingqiao Liu, Ian Reid, Peter Corke},pdftitle={Exploiting Temporal Information for DCNN-based Fine-Grained Object Classification}]{hyperref}
\usepackage{balance}
\usepackage[absolute]{textpos}

\renewcommand\tablename{\normalfont\bf{Table}}

\makeatletter
\def\blfootnote{\gdef\@thefnmark{}\@footnotetext}
\makeatother

\begin{document}

\title{\scalebox{1.4}{\normalsize\bf Exploiting Temporal Information for DCNN-based Fine-Grained Object Classification}}

\author
  {
  ZongYuan Ge, Chris McCool, Conrad Sanderson, Peng Wang, Lingqiao Liu, Ian Reid, Peter Corke\\
  \small
  ~\\
  Australian Centre for Robotic Vision, Australia\\
  Queensland University of Technology, Australia\\
  Data61, CSIRO, Australia\\
  University of Queensland, Australia\\
  University of Adelaide, Australia
  }

\maketitle

\begin{abstract}

Fine-grained classification is a relatively new field that has concentrated on using information from a single image,
while ignoring the enormous potential of using video data to improve classification.
In this work we present the novel task of video-based fine-grained object classification,
propose a corresponding new video dataset,
and perform a systematic study of several recent deep convolutional neural network (DCNN) based approaches, which we specifically adapt to the task.
We evaluate three-dimensional DCNNs, two-stream DCNNs, and bilinear DCNNs.
Two forms of the two-stream approach are used, where spatial and temporal data from two independent DCNNs are fused
either via early fusion (combination of the fully-connected layers) and late fusion (concatenation of the softmax outputs of the DCNNs).
For bilinear DCNNs, information from the convolutional layers of the spatial and temporal DCNNs is combined via local co-occurrences.
We then fuse the bilinear DCNN and early fusion of the two-stream approach to combine the spatial and temporal information at the local and global level (Spatio-Temporal Co-occurrence).
Using the new and challenging video dataset of birds,
classification performance is improved from 23.1\% (using single images) to 41.1\% when using the Spatio-Temporal Co-occurrence system.
Incorporating automatically detected bounding box location further improves the classification accuracy to 53.6\%.
\end{abstract}

\blfootnote{\hspace{-2.3ex}\scalebox{0.725}{\textbf{Published in:} International Conference on Digital Image Computing: Techniques and Applications, 2016.}\\\scalebox{0.75}{Associated video dataset is available at: \href{http://arma.sf.net/vb100/}{\bf http://arma.sf.net/vb100/}}}

\vspace{-2ex}
\section{Introduction}

Fine-grained object classification consists of discriminating between classes in a sub-category of objects,
for instance the particular species of bird or dog~\cite{berg2013poof,chai2013symbiotic,farrell2011birdlets,gavves2013fine,zhang2014part}.
This is a very challenging problem due to large intra-class variations caused by pose and appearance changes,
as well as small inter-class variation due to subtle differences in the overall appearance between classes~\cite{berg2013you,ge2015modelling}.

Prior work in fine-grained classification has concentrated on learning image-based features to cope with pose variations.
Initially such approaches used traditional image-based features such as colour and histograms of gradients~\cite{berg2013poof}
while modelling the pose using a range of methods including deformable parts-based approaches~\cite{chai2013symbiotic,liu2012dog,zhang2013deformable}.
More recently, deep convolutional neural networks (DCNNs) have been used to learn robust features~\cite{donahue2013decaf},
cope with large variations by using a hierarchical model~\cite{ge2016fine},
and automatically localise regions of importance~\cite{jaderberg2015spatial}.
Despite the advances provided by these approaches,
prior work treats the fine-grained classification task as a still-image classification problem and ignores complementary temporal information present in videos.

Recent work on neural network based approaches has provided notable results in video-based recognition~\cite{ge2015modelling,karpathy2014large,simonyan2014two,tran2014learning,yue2015beyond}.
Karpathy et al.~\cite{karpathy2014large} demonstrated the surprising result that classifying a single frame from a video using a DCNN
was sufficient to perform accurate video classification, for broad categories such as activity and sport recognition.
Within the action recognition area, 
Simonyan and Zisserman~\cite{simonyan2014two} incorporate optical flow and RGB colour information into two stream networks.
Tran~et~al.~\cite{tran2014learning} apply deep 3D convolutional networks (3D ConvNets)
to implicitly learn motion features from raw frames and then aggregate predictions at the video level.
Ng~et~al.~\cite{yue2015beyond} employ Long Short-Term Memory cells which are connected to the output of the underlying CNN
to achieve notable results on the UCF-101~\cite{soomro2012ucf101} and Sports 1 million datasets~\cite{karpathy2014large}. 
To date, the above neural network based approaches have not been explored for the task of video-based fine-grained object classification.

\textbf{Contributions.}
In this paper, we introduce the problem of video-based fine-grained object classification,
propose a corresponding new dataset,
and explore several methods to exploit the temporal information.
A~systematic study is performed comparing several DCNN based approaches which we have specifically adapted to the task,
highlighting the potential benefits that fine-grained object classification can gain by modelling temporal information.
We evaluate 3D~ConvNets~\cite{tran2014learning},
two-stream DCNNs~\cite{simonyan2014two},
and bilinear DCNNs~\cite{lin2015bilinear}.
Two forms of the two-stream approach are used:
(i) the originally proposed late-fusion form which concatenates the softmax outputs of two independent spatial and temporal DCNNs,
and
(ii) our modified form, which performs early-fusion via combination of the fully-connected layers.
In contrast to the two forms of the two-stream approach, we adapt the bilinear DCNN to extract local co-occurrences
by combining information from the convolutional layers of spatial and temporal DCNNs.
The adapted bilinear DCNN is then fused with the two-stream approach (early fusion) to combine spatial and temporal information at the local and global level.

The study is performed on the VB100 dataset, a new and challenging video dataset of birds
consisting of 1,416 video clips of 100 species birds taken by expert bird watchers.
The dataset contains several compounded challenges, such as clutter, large variations in scale, camera movement and considerable pose variations.
Experiments show that classification performance is improved from 23.1\% (using single images) to 41.1\% when using the spatio-temporal bilinear DCNN approach,
which outperforms 3D~ConvNets as well as both forms of the two-stream approach.
We highlight the importance of performing early fusion, either at the input layer (3D~ConvNets) or feature layer (adapted bilinear DCNN),
as this consistently outperforms late fusion (ie.~the original two-stream approach).
Incorporating automatically detected bounding box location further improves the classification accuracy of the spatio-temporal bilinear DCNN approach to 53.6\%.

We continue the paper as follows.
Section~\ref{sec:method} describes the studied methods and our adaptations,
while Section~\ref{sec:dataset} describes the new VB100 bird dataset.
Section~\ref{sec:experiments} is devoted to comparative evaluations.
The main findings are summarised in Section~\ref{sec:conclusion}.

\section{Combining Spatial and Temporal Information}
\label{sec:method}

In this section we first describe two baseline networks that make use of either image or temporal information.
We then outline the deep 3-dimensional convolutional network~\cite{tran2014learning},
extend the two-stream approach~\cite{simonyan2014two}
and adapt the bilinear DCNN approach~\cite{lin2015bilinear} to encode local spatial and temporal co-occurrences.

\subsection{Underlying Spatial and Temporal Networks}

Our baseline systems are DCNNs that use as input either optical flow (temporal) or image-based features.
The temporal network $\mathcal{T}$ uses as input
the horizontal flow $\mathbf{O}_{x}$, vertical flow $\mathbf{O}_{y}$, and magnitude of the optical flow $\mathbf{O}_{mag}$
combined to form a single optical feature map $\mathbf{O}\in\mathrm{R}^{h \times w \times 3}$,
where $h \times w$ is the size of the feature map (image).
The spatial network $\mathcal{S}$ uses RGB frames (images) as input.
Both $\mathcal{S}$ and $\mathcal{T}$ use the DCNN architecture of Krizhevsky et al.~\cite{krizhevsky2012imagenet}
which consists of 5 convolutional layers, $\mathbf{S}^{c1}, \mathbf{S}^{c2}, \dots, \mathbf{S}^{c5}$,
followed by 2 fully connected layers, $\mathbf{S}^{fc6}$ and $\mathbf{S}^{fc7}$,
prior to the softmax classification layer, $\mathbf{S}^{o}$.
The networks are trained by considering each input frame from a video (either image or optical flow) to be a separate instance,
and are fine-tuned to the specific task (and modality) by using a pre-trained network.
Fine-tuning~\cite{yosinski2014transferable} is necessary as we have insufficient classes and observations to train the networks from scratch
(preliminary experiments indicated that training the networks from scratch resulted in considerably lower performance).

When performing classification, each image (or frame of optical flow) is initially treated as an independent observation.
For a video of $N_{f}$ frames this leads to $N_{f}$ classification decisions.
To combine the decisions, the max vote of these decisions is taken.

\subsection{Deep 3D Convolutional Network}

The deep 3-dimensional convolutional network (3D~ConvNet) approach~\cite{tran2014learning},
originally proposed for action recognition,
utilises 3-dimensional convolutional kernels to model $L$ frames of information simultaneously.
In contrast to optical flow features where temporal information is explicitly modelled,
the approach implicitly models the information within the deep neural network structure.
This approach obtains state-of-the-art performance on various action recognition datasets
such as \mbox{UCF-101}~\cite{soomro2012ucf101} and ASLAN~\cite{kliper2012action}. 
The network is fine-tuned for our classification task by taking a sliding window of $L=15$ frames and moving the sliding window one frame at a time; each sliding window is considered to be a separate instance.
This results in $N_{f} - 14$ classification decisions which are combined using the max vote.

\subsection{Spatio-Temporal Two-Stream Network: Early and Late Fusion}
\label{sec:two_stream_forms}

\begin{figure}[!b]
  \centering
  \includegraphics[width=\columnwidth]{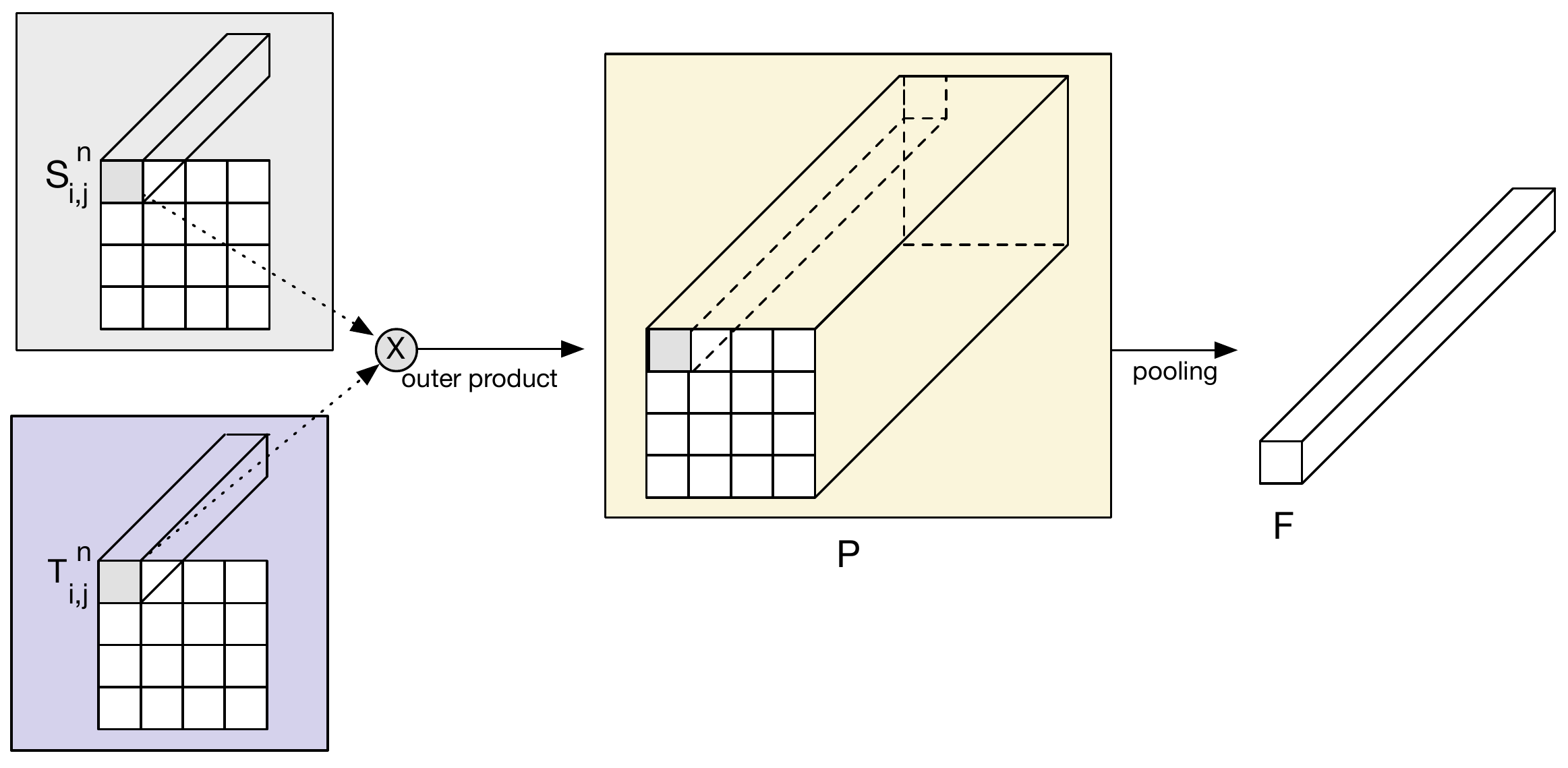}
  \caption
    {
    Conceptual illustration of the spatio-temporal co-occurrence approach.
    }
  \label{fig:pool}
\end{figure}

The two-stream network proposed for action recognition by Simonyan and Zisserman~\cite{simonyan2014two}
uses the two independent spatial and temporal networks $\mathcal{S}$ and $\mathcal{T}$.
The softmax output of these two networks 
is then concatenated and used as a feature vector that is classified by a multi-class support vector machine (SVM).
We refer to this network as {\it Two-Stream (late fusion)}; it is conceptually illustrated in Fig.~\ref{fig:key_figure}(a).

A potential downside of this approach is that fusion of spatial and temporal information is done at the very end.
This limits the amount of complementary information captured as scores (or decisions) from the softmax classification layer are combined.
To address this issue, we propose to combine the two streams of information much earlier (early fusion)
by combining the {\it fc6} outputs, $\mathbf{S}^{fc6}$ and $\mathbf{T}^{fc6}$;
{\it fc6} is the first fully connected layer and is often used to extract a single feature from DCNNs~\cite{donahue2013decaf}.
We refer to this modified network as {\it Two-Stream (early fusion)}.
See Fig.~\ref{fig:key_figure}(b).

\subsection{Joint Spatial and Temporal Features via Co-occurrences}

We adapt the recently proposed bilinear DCNN approach by Lin et al.~\cite{lin2015bilinear}
via combining the convolutional layers of the baseline spatial and temporal networks 
by calculating co-occurrences. The rationale behind is that different species of birds may have different appearance and motion patterns and their combination.
Specifically, 
let the feature maps of the $n$-th layer of the spatial and temporal networks be
$\mathbf{S}^n\in\mathbb{R}^{h\times{w}\times{d_n}}$ and $\mathbf{T}^n\in\mathbb{R}^{h\times{w}\times{d_n}}$,
where $d_n$ is the number of dimensions for the feature map (number of kernels). 
The two feature maps are combined by calculating an outer product:

\begin{equation}
\mathbf{P}_{i,j} = \operatorname{vec} \left( {\mathbf{S}^n_{i,j}}{\mathbf{T}^n_{i,j}}^{\intercal} \right)
\label{eqn:outer_product}
\end{equation}

\noindent
where $\mathbf{S}^n_{i,j}\in\mathbb{R}^{d_n}$ and $\mathbf{T}^n_{i,j}\in\mathbb{R}^{d_n}$
are the local feature vectors of the spatial and temporal streams at location $(i,j)$,
$\operatorname{vec(\cdot)}$ is the vectorisation operation,
and
$\mathbf{P} \in\mathbb{R}^{h \times w \times d^2_n}$, with $\mathbf{P}_{i,j} \in \mathbb{R}^{d^2_n}$ being the co-occurrence feature at location $(i,j)$.
As such, the outer product operation captures the co-occurrence of the visual and motion patterns at each spatial location. Max pooling is applied to all the local encoding vectors $\mathbf{P}_{i,j}$ to create the final feature representation $\mathbf{F}\in\mathbb{R}^{d_{n}^2}$.
Finally, $L_2$ normalisation is applied to the encoding vector~\cite{lin2015bilinear}.
The overall process is conceptually illustrated in Fig.~\ref{fig:pool}. 

\begin{figure}[!t]
  \centering
  \begin{minipage}{1\columnwidth}
  \centering
    \begin{minipage}{0.80\textwidth}
      \centering
      \includegraphics[width=1\textwidth]{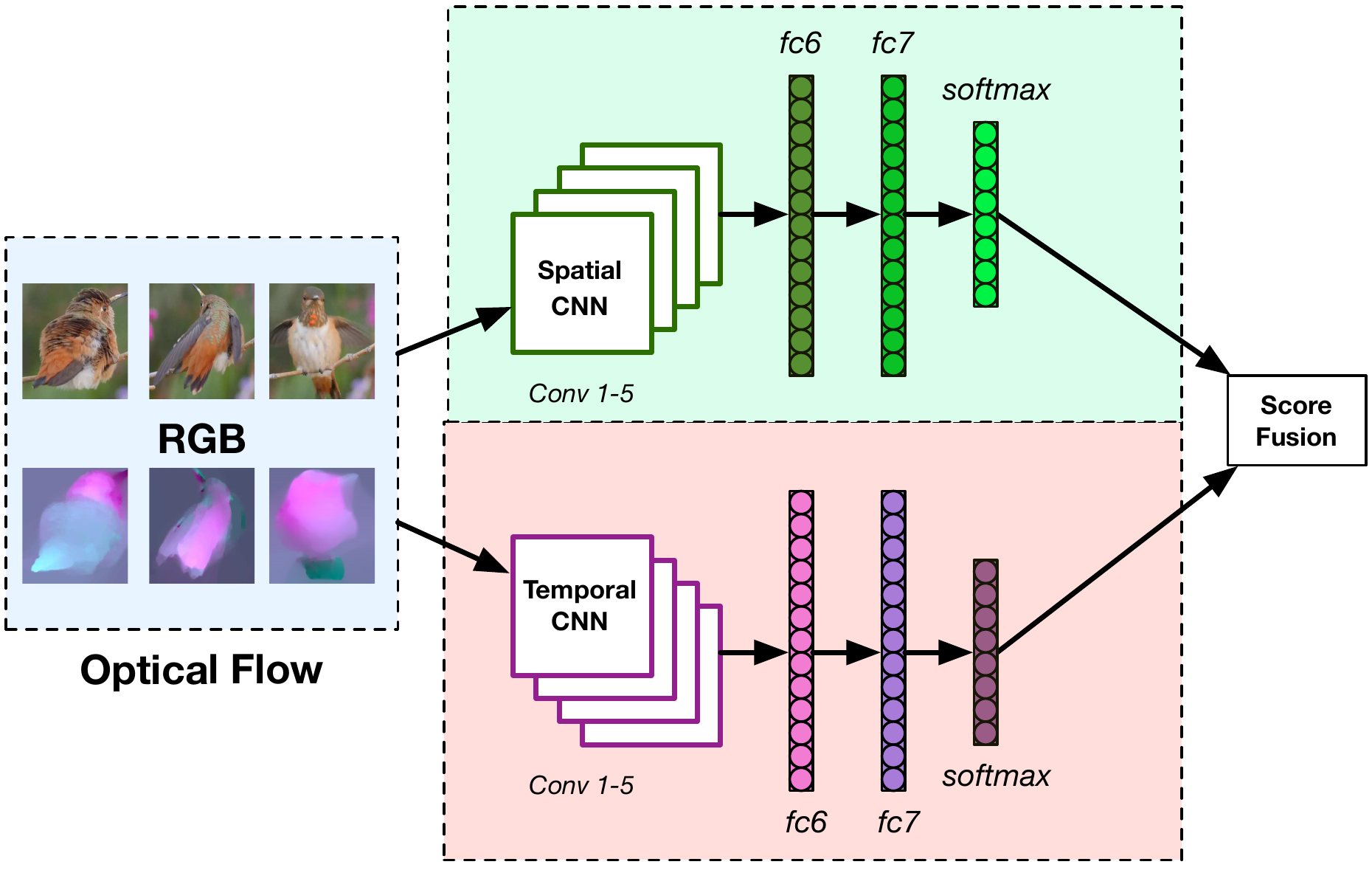}\\
      ~~~~~~(a) Two-Stream (late fusion)
    \end{minipage}
    
    \vspace{4ex}
    
    \begin{minipage}{0.90\textwidth}
      \centering
      ~~~~~~\includegraphics[width=1\textwidth]{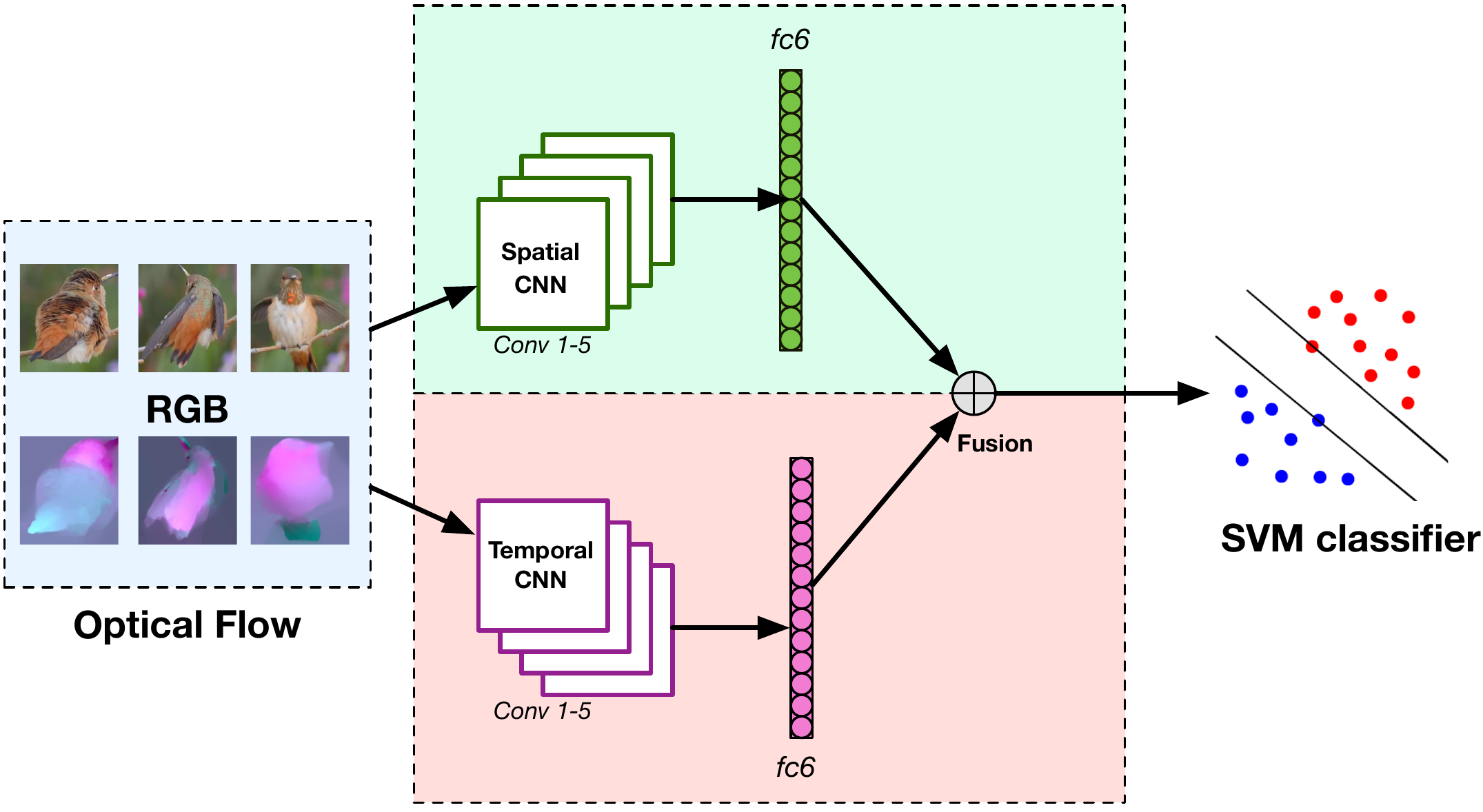}\\
      ~~~~~~(b) Two-Stream (early fusion)
    \end{minipage}
    
    \vspace{4ex}
    
    \begin{minipage}{0.90\textwidth}
      \centering
      ~~~~~~~\includegraphics[width=1\textwidth]{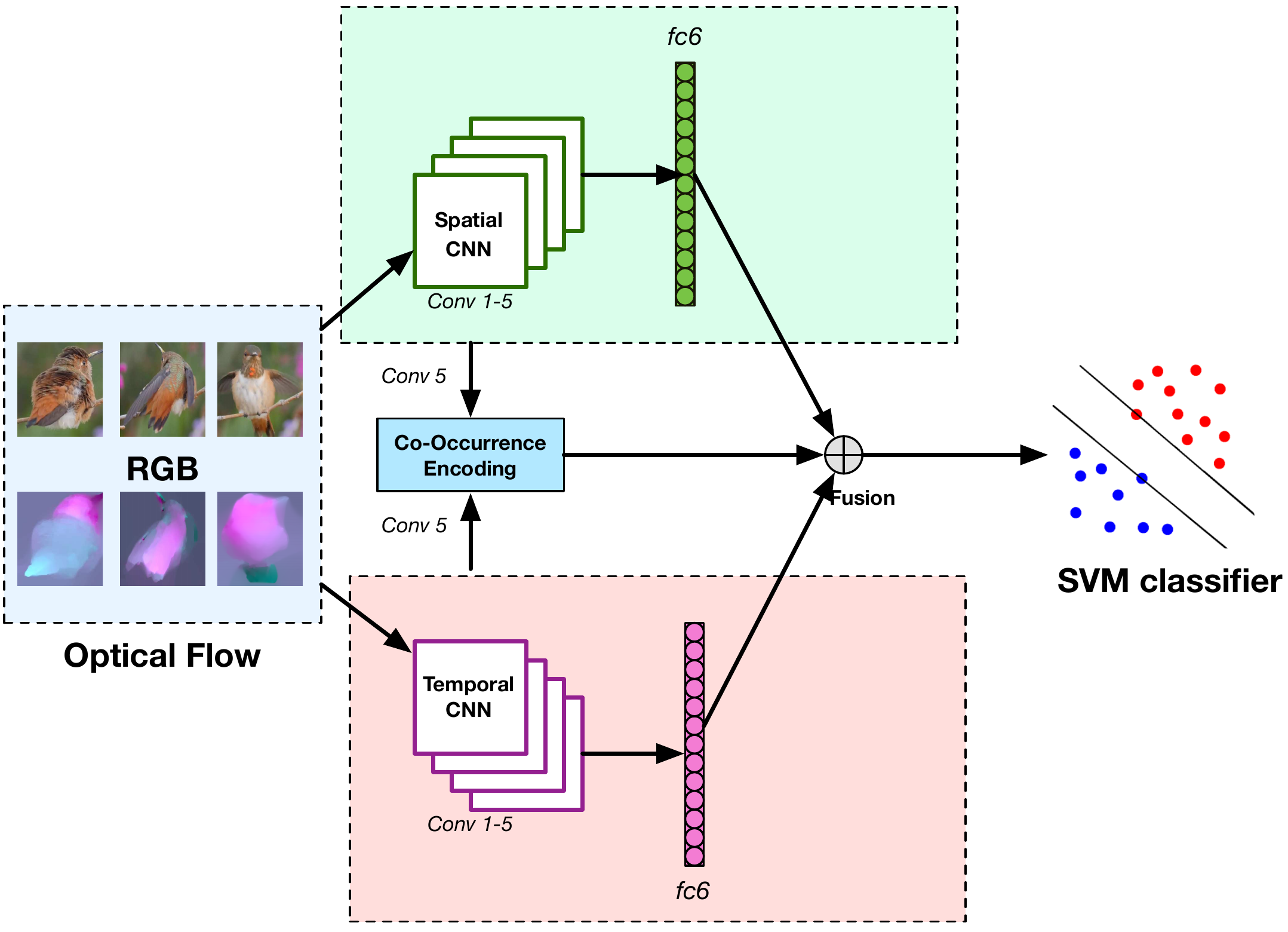}\\
      ~~~~~~(c) Spatio-Temporal Co-Occurrence
    \end{minipage}
    \vspace{1ex}

  \end{minipage}
  
  \caption
    {
    Overview of the Two-Stream and Spatio-Temporal Co-Occurrence approaches for fine-grained video classification.
    In (a) the Two-Stream approach uses {\it late fusion}, where features are combined from the softmax layer.
    In (b) the Two-Stream approach uses {\it early fusion}, where features are combined from the {\it fc6} layer.
    The Spatio-Temporal Co-Occurrence approach (c) combines the co-occurrence (bilinear DCNN) features with the features from {\it fc6}.
    }
  \label{fig:key_figure}
\end{figure}

The spatio-temporal bilinear DCNN feature is combined with the {\it fc6} spatial and temporal features used for {\it Two-Stream (early fusion)}.
This allows us to combine the spatial and temporal information at both the local and global level.
The resultant features are fed to an SVM classifier.
See Fig.~\ref{fig:key_figure}(c) for a conceptual illustration.
We refer this system as {\it Spatio-Temporal Co-occurrence}.

\newpage
\textcolor{white}{.}
\newpage
\section{VB100 Dataset: Videos of 100 Bird Species}
\label{sec:dataset}
\vspace{1ex}

\begin{figure}[!t]
  \centering
  
  \vspace{1.5ex}
  
  \begin{minipage}{1\columnwidth}
  \centering
  \includegraphics[width=0.32\columnwidth]{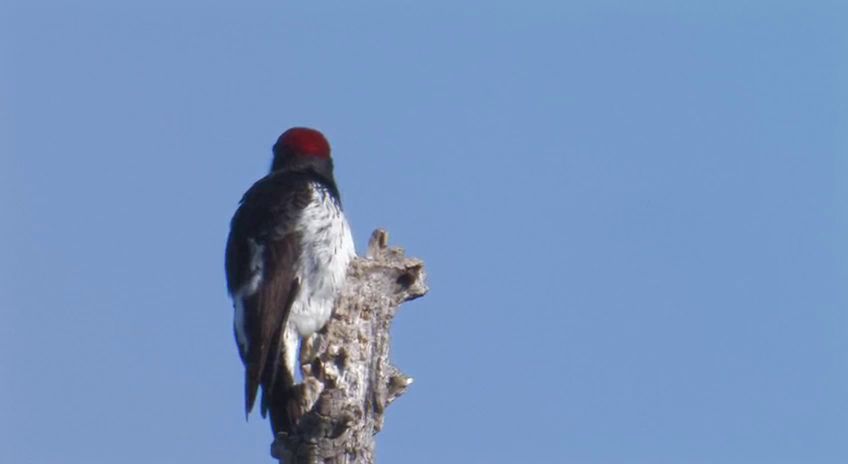}
  \includegraphics[width=0.32\columnwidth]{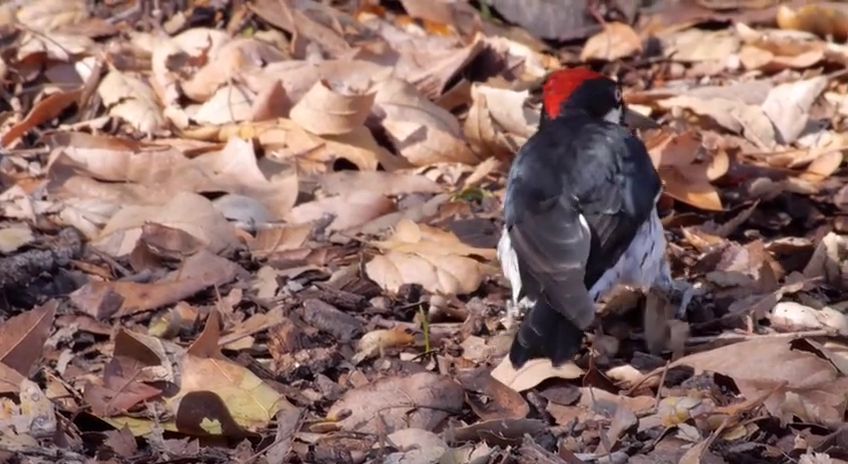}
  \includegraphics[width=0.32\columnwidth]{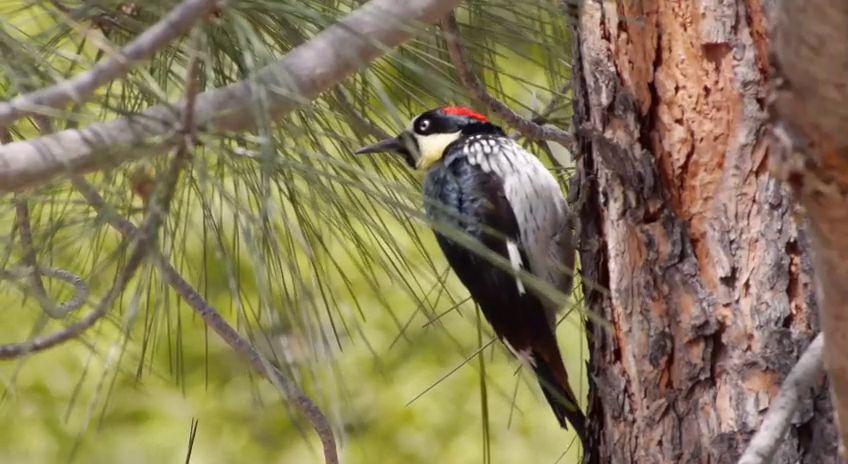}
  \end{minipage}
  
  \vspace{1.5ex}
  
  \begin{minipage}{1\columnwidth}
  \centering
  \includegraphics[width=0.32\columnwidth]{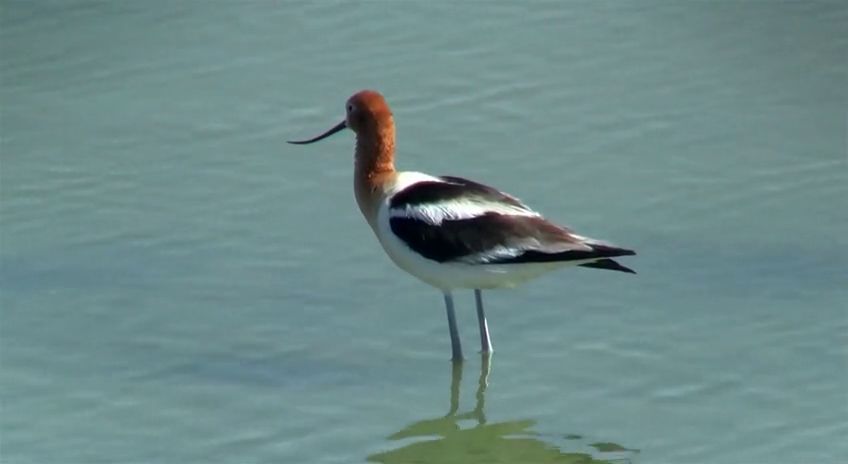}
  \includegraphics[width=0.32\columnwidth]{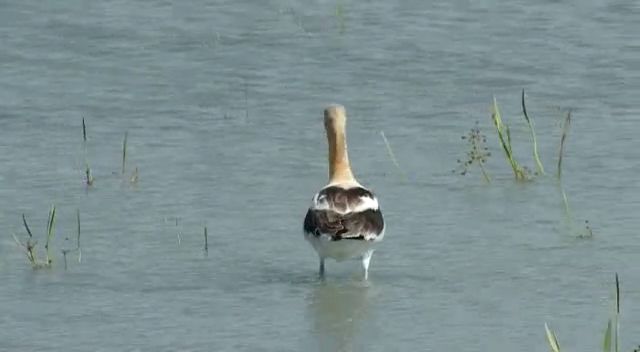}
  \includegraphics[width=0.32\columnwidth]{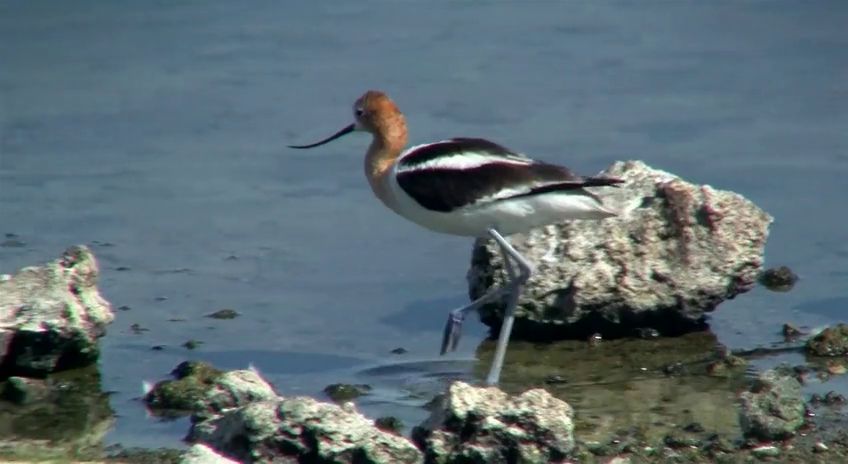}
  \end{minipage}
  
  \vspace{1.5ex}
  
  \begin{minipage}{1\columnwidth}
  \centering
  \includegraphics[width=0.32\columnwidth]{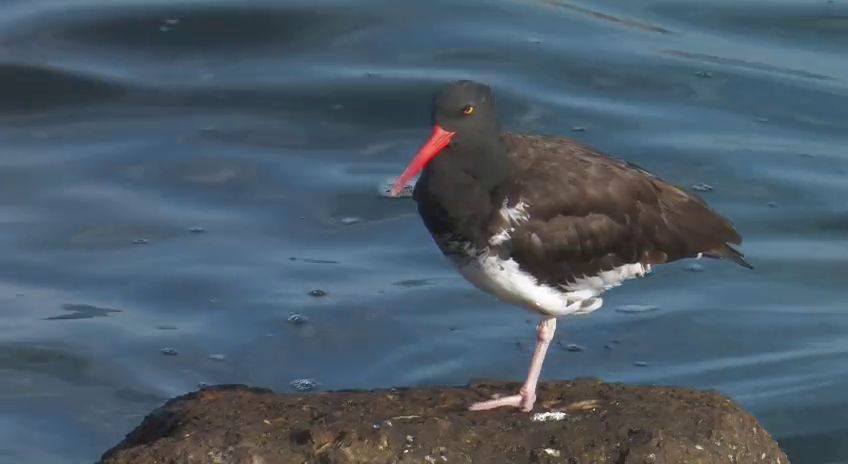}
  \includegraphics[width=0.32\columnwidth]{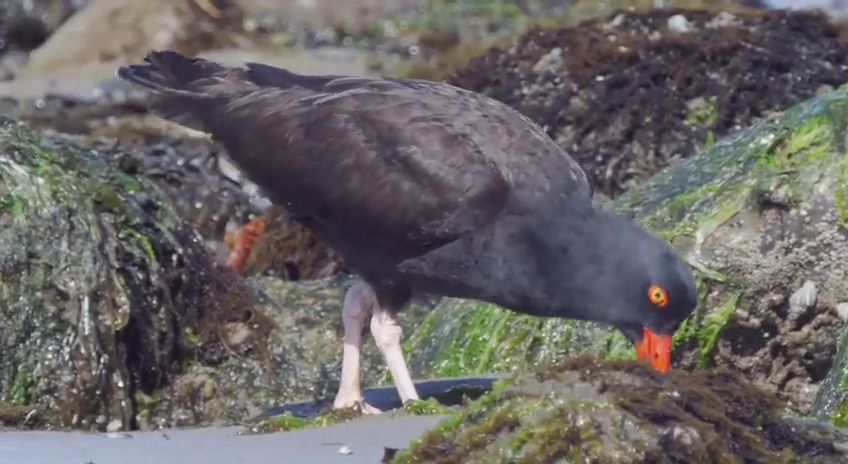}
  \includegraphics[width=0.32\columnwidth]{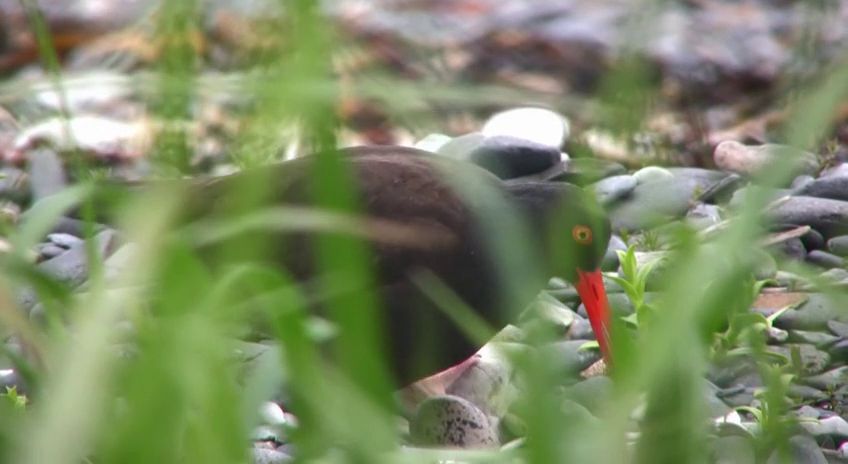}
  \end{minipage}
  
  \caption
    {
    \small
    Example frames from video clips in the VB100 dataset.
    Each row shows three sample frames for a unique class.
    The first frame in each row (left to right) shows an easy situation, followed by images with variations such as pose, scale and background. 
    }
  \label{fig:dataset}
\end{figure}

To investigate video-based fine-grained object classification we propose the VB100 dataset,
a new and challenging dataset consisting of 1,416 video clips of 100 bird species taken by expert bird watchers.
The birds were often recorded at a distance, introducing several challenges such as large variations in scale, bird movement, camera movement and considerable pose variations.
See Fig.~\ref{fig:dataset} for examples.

\begin{figure}[!t]
  \vspace{2ex}
  \centering
  \includegraphics[width=1\columnwidth]{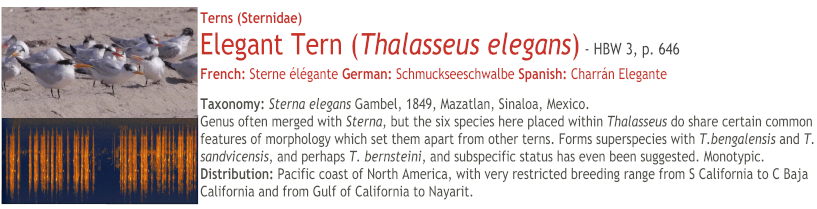}\\
  \vspace{-1ex}
  \caption
    {
    \small
    An example for the class {\it Elegant Tern} in VB100.
    Top-left: a~still shot from one of the video clips.
    Bottom-left: spectrogram created from the corresponding audio file.
    Right: taxonomy information.
    }
  \label{fig:sample}
\end{figure}

For each class (species of bird), the following data is provided: 
video clips, sound clips, as well as taxonomy and distribution location.
See Fig.~\ref{fig:sample} for an example.

Each class has on average 14 video clips.
The median length of a video is 32 seconds.
The frame rate varies across the videos; approximately 69\% of videos were captured at 30 frames per second (fps), 30\% at 25 fps, and the remaining at 60 and 100 fps.

Often the camera will need to move in order to track the bird, keeping it in view;
this form of camera movement is present in 798 videos, with the remaining 618 videos obtained using either static or largely static cameras.

The dataset can be obtained from:
\href{http://arma.sf.net/vb100/}{\small\textsf{http://arma.sf.net/vb100/}}

\newpage
\section{Experiments}
\label{sec:experiments}

Two sets of experiments are presented in this section.
In the first set (Section~\ref{sec:comparative_evaluation}), we evaluate the performance without taking into account whether each video clip was recorded by a static or moving camera.
In the second set (Section~\ref{sec:static_vs_moving}), we study the effect of camera movement on performance.
In all cases, to obtain a per video classification decision we use the max voting from the classified frames.
For the Spatio-Temporal Co-occurrence approach, initial experiments found that using the last convolutional layer $n=c5$ provided the best performance;
this leads to $d=65,536$ for the spatio-temporal bilinear features.
The input frame size for all networks is \mbox{$224\times224$}.
Training and testing is performed using Caffe~\cite{jia2014caffe}.

The dataset is divided into 730 training videos (train set) and 686 testing videos (test set).
Results are presented in terms of mean classification accuracy.
Classification accuracy is calculated on a per video basis and per class basis,
with $\mbox{accuracy} = {N^{c}_{p}} / {N^{c}}$,
where $N^{c}_{p}$ is the number of correctly classified videos for the $c$-th class and $N^{c}$ is the number of videos for the $c$-th class.
The mean classification accuracy is then calculated across all of the classes.

\subsection{Comparative Evaluation}
\label{sec:comparative_evaluation}

\begin{figure}[!b]
  \centering
  \begin{minipage}{1\columnwidth}
    \begin{minipage}{1\textwidth}
    \centering
      \includegraphics[width=0.32\textwidth]{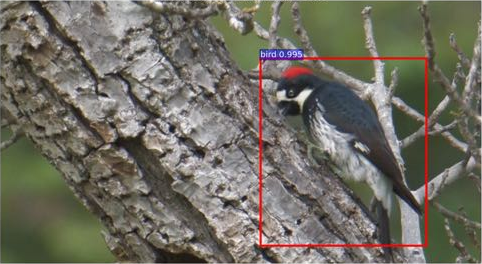} 
      \includegraphics[width=0.32\textwidth]{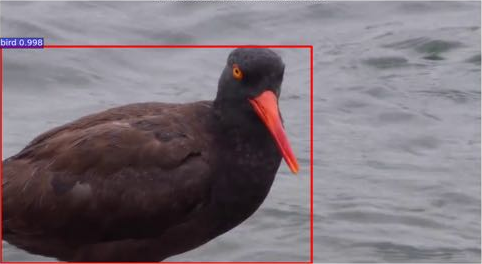}
      \includegraphics[width=0.32\textwidth]{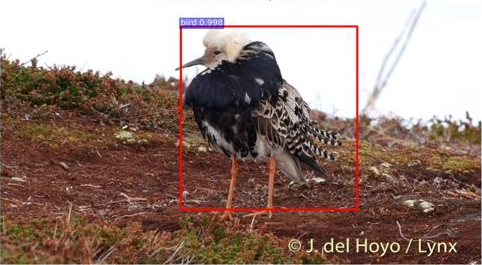}
    \end{minipage}
    
    ~\\
    
    \begin{minipage}{1\textwidth}
    \centering
      \includegraphics[width=0.32\textwidth]{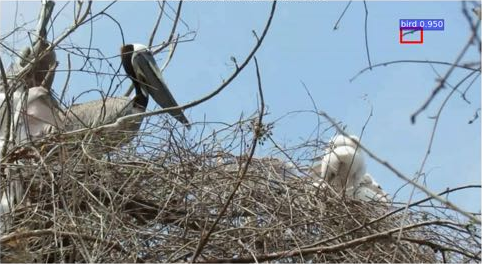} 
      \includegraphics[width=0.32\textwidth]{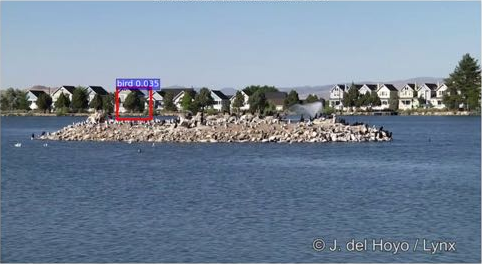}
      \includegraphics[width=0.32\textwidth]{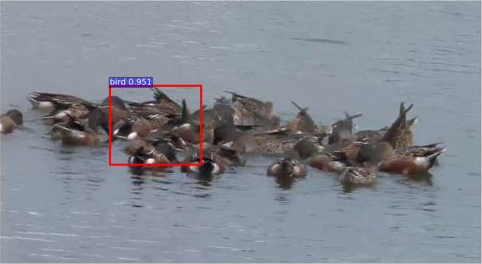}
    \end{minipage}
  \end{minipage}
  \caption
    {
    Examples of bird localisation (red bounding box) using the default settings of Faster R-CNN~\cite{ren2015faster}.
    Top row: good localisations.
    Bottom row: bad localisations due to confounding textures, clutter, small objects, and occlusions.
    }
  \label{fig:detect}
\end{figure}

We first investigate the performance of two independent networks for spatial and temporal information: Spatial-DCNN and Temporal-DCNN. 
We then compare the performance of 
3D~ConvNets~\cite{tran2014learning} fine-tuned for our bird classification task (referred to as 3D~ConvNets-FT),
the two-stream approach~\cite{simonyan2014two} (which combines the Spatial-DCNN and Temporal-DCNN networks),
and the spatio-temporal co-occurrence approach.
Finally we evaluate the performance of the co-occurrence approach in conjunction with an off-the-shelf bird detector/locator.
For this we use the recent Faster Region CNN~\cite{ren2015faster} approach 
with default parameters learned for the PASCAL VOC challenge~\cite{everingham2010pascal};
only bird localisations are used, with all other objects ignored.
Examples of localisation are shown in Fig.~\ref{fig:detect}.

\textbf{Network Setup.}
The Spatial-DCNN uses the AlexNet structure pre-trained on the ImageNet dataset~\cite{krizhevsky2012imagenet} before being fine-tuned for our bird classification task.
It is trained by considering each frame from a video to be a separate instance (image).
Two variants of Spatial-DCNN are used:
(i)~randomly selecting one frame per video clip,
and
(ii)~using 5 frames per second (fps) from each video clip\footnote{The video clips were normalised to 5 fps, as this was computationally more efficient. Preliminary experiments indicated that using 5 fps leads to similar performance as normalising at 25 fps.}.
The Temporal-DCNN uses dense optical flow features computed from the Matlab implementation of Brox et al.~\cite{brox2004high}. 
For the sake of computational efficiency, we have calculated the optical flow every 5 frames.

It is generally beneficial to perform zero-centering of the network input,
as it allows the model to better exploit the rectification non-linearities and for optical flow features provides robustness to camera movement~\cite{simonyan2014two}. 
Therefore, for both Spatial-DCNN and Temporal-DCNN we perform mean normalisation of the input data.
For Spatial-DCNN we subtract the mean value for each RGB channel,
while for Temporal-DCNN mean flow subtraction is performed for the temporal input. 

For the two-stream approach we use two forms (as described in Section~\ref{sec:two_stream_forms}):
(i)~early fusion, where the first fully connected features (fc6) from the Spatial-DCNN (with 5 fps) and Temporal-DCNN networks are concatenated,
and
(ii)~late fusion, where the softmax output of the two networks is concatenated.
For the two-stream and the spatio-temporal co-occurrence approaches,
the resultant feature vectors are fed to a multi-class linear SVM for classification.

\textbf{Quantitative Results.}
The results presented in Table~\ref{table:baseline_results}
show that using more frames from each video (ie. more spatial data) leads to a notable increase in accuracy.
This supports the use of videos for fine-grained classification.
The results also show that spatial data provides considerably more discriminatory information than temporal data.
In all cases, combining spatial and temporal information results in higher accuracy than using either type of information alone,
confirming that the two streams of data carry some complementary information.

In contrast to the using late fusion in the standard two-stream approach,
performing early fusion yields a minor increase in accuracy ($37.5\%$ vs $38.9\%$) and slightly exceeds the accuracy obtained by 3D~ConvNets-FT ($38.6\%$).
Using the co-occurrence approach leads to the highest fusion accuracy of $41.1\%$.
This highlights the importance of making use of the extra information from the video domain for object classification.
Finally, using the Spatio-Temporal Co-occurrence system in conjunction with an automatic bird locator increases the accuracy from $41.1\%$ to $53.6\%$.
This in turn highlights the usefulness of focusing attention on the object of interest and reducing the effect of nuisance variations.

\begin{table}[!b]
\small
\setlength{\tabcolsep}{4pt}
\centering
\caption{Fine-grained video classification results on the VB100 video dataset.}
\label{table:baseline_results}
\begin{tabular}{lc}
\hline\hline\noalign{\smallskip}
{\bf Method} & {\bf Mean Accuracy}\\
\noalign{\smallskip}
\hline\hline
\noalign{\smallskip}
Spatial-DCNN (random frame)                   & 23.1\% \\  
Spatial-DCNN (5 fps)                          & 37.0\% \\
Temporal-DCNN ($\Delta=5$)                    & 22.9\% \\ \hline
Two-Stream (early fusion)                     & 38.9\% \\
Two-Stream (late fusion)                      & 37.5\% \\
3D~ConvNets-FT                                & 38.6\% \\ \hline
Bilinear DCNNs~\cite{lin2015bilinear}         & 33.8\% \\
Spatio-Temporal Co-occurrence                  & 41.1\% \\
Spatio-Temporal Co-occurrence + bounding box   & 53.6\% \\
\hline
\end{tabular}
\end{table}

\begin{figure}[!b]
  \centering
  \vline
  \begin{minipage}{1\columnwidth}
    \centering
    \hrule
    \begin{minipage}{0.45\textwidth}
      \centering
      \includegraphics[width=1\textwidth]{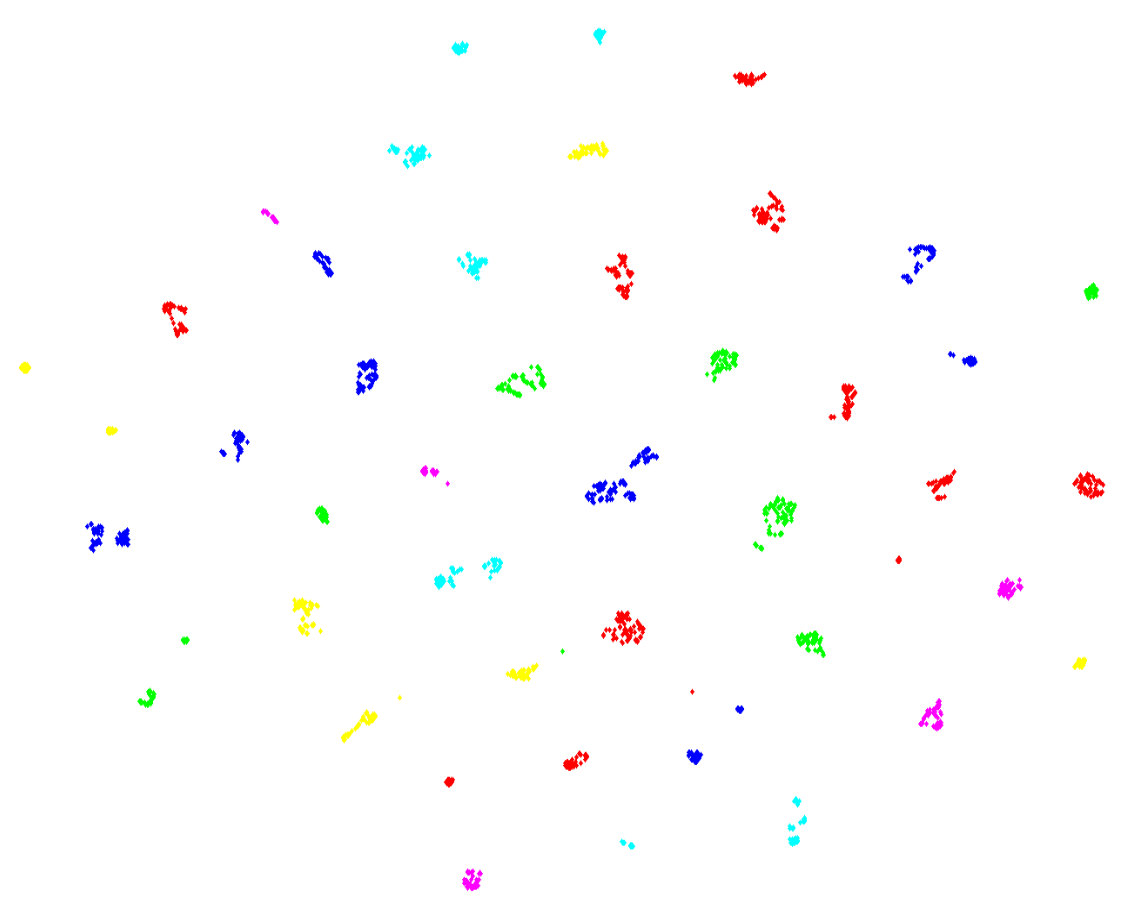}
    \end{minipage}
    \hfill
    \vline
    \hfill
    \begin{minipage}{0.45\textwidth}
      \centering
      \includegraphics[width=1\textwidth]{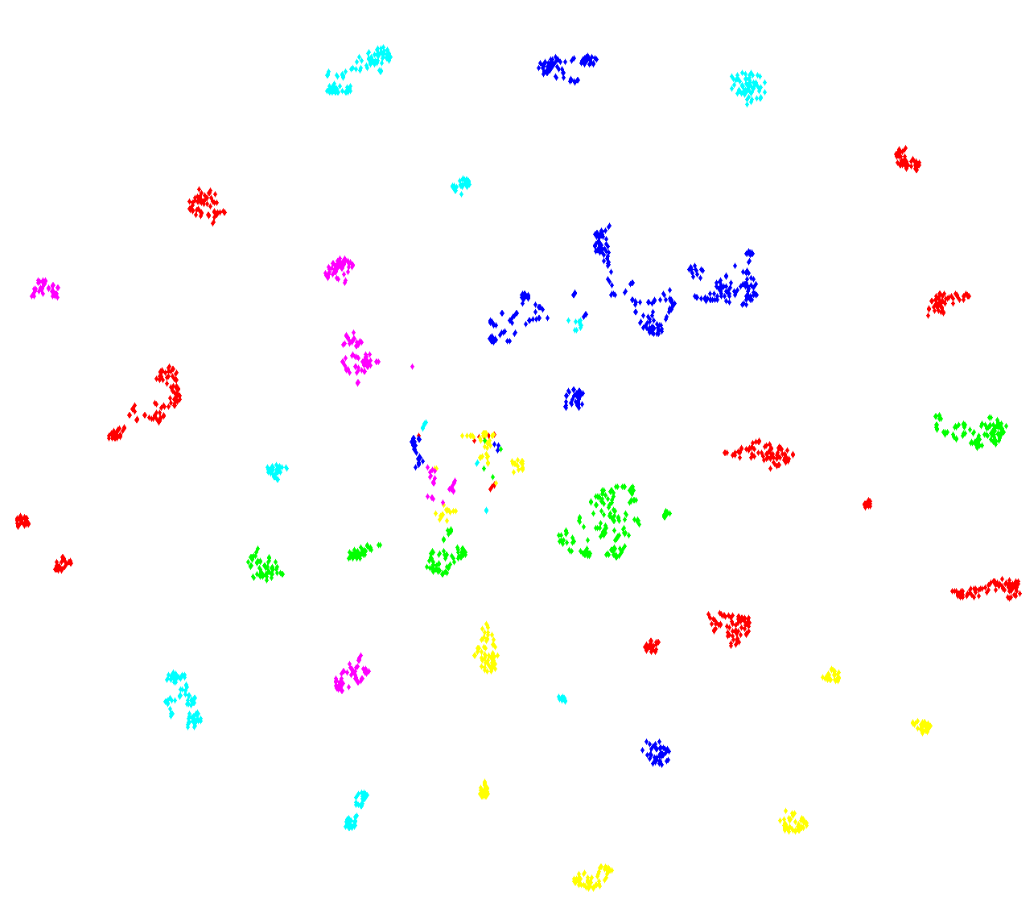}
    \end{minipage}
  \vline
  \hrule
  \end{minipage}
  
  \begin{minipage}{1\columnwidth}
    \centering
    \begin{minipage}{0.45\textwidth}
      \centering
      ~\\
      {\small (a)}
    \end{minipage}
    \begin{minipage}{0.45\textwidth}
      \centering
      ~\\
      {\small (b)}
    \end{minipage}
  \end{minipage}

  \caption
    {
    Qualitative evaluation using t-SNE~\cite{van2008visualizing} to visualise the data for 10 classes indicated by unique colours:
    (a)~using Spatial-DCNN features,
    and
    (b)~using Spatio-Temporal Co-occurrence features.
    For both approaches several distinct clusters are formed for each class.
    By using the co-occurrence approach fewer separated clusters are formed, and the separated clusters tend to be closer together.
    }
  \label{fig:qualitative_eval}
\end{figure}

\textbf{Qualitative Results.}
To further examine the impact of incorporating temporal information via the co-occurrence approach, 
we visualise 10 classes with features taken from the Spatial-DCNN and Spatio-Temporal Co-occurrence approaches.
To that end we use the t-Distributed Stochastic Neighbour Embedding (t-SNE) data visualisation technique based on dimensionality reduction~\cite{van2008visualizing}.
In Fig.~\ref{fig:qualitative_eval} it can be seen that both sets of features yields several distinct clusters for each class.
However, by using the co-occurrence approach fewer separated clusters are formed,
and the separated clusters tend to be closer together.
This further indicates that benefit can be obtained from exploiting temporal information in addition to spatial information.

\subsection{Effect of Camera Type: Static vs Moving}
\label{sec:static_vs_moving}

\begin{figure}[!b]
  \centering
  \begin{minipage}{1\columnwidth}
  \centering
    \includegraphics[width=0.32\textwidth]{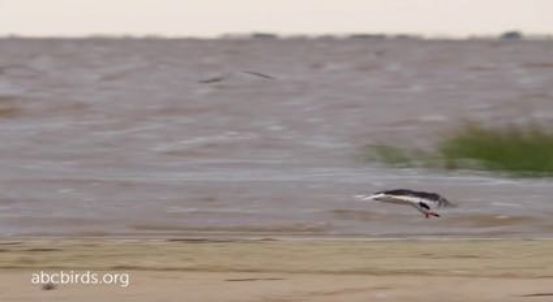} 
    \includegraphics[width=0.32\textwidth]{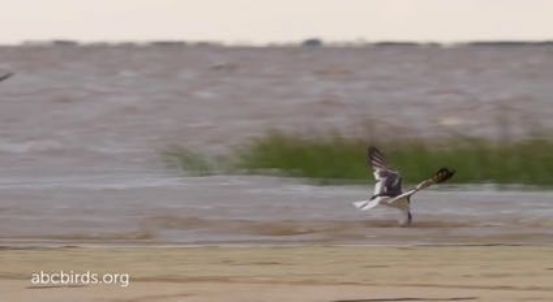}
    \includegraphics[width=0.32\textwidth]{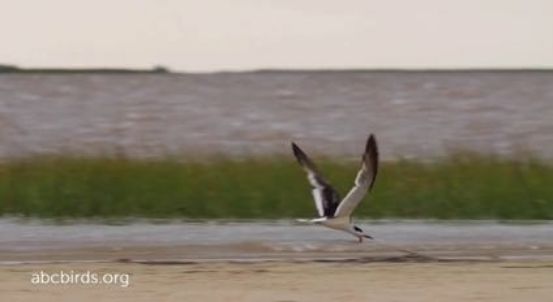}
  \end{minipage}
  \caption
    {
    Examples of video frames recorded by a moving camera, manually tracking the bird.
    }
  \label{fig:static}
\end{figure}

In this section we explore how camera motion affects performance. 
Camera motion is a dominant variation within the VB100 dataset as it contains 618 video clips recorded with a static camera and 798 video clips recorded with a moving camera, which follow bird movement (eg., flight).
Fig.~\ref{fig:static} shows examples from two videos of Elegant Tern recorded by static and moving cameras.

Previous work in action recognition~\cite{Jain13_1:conference,Kuehne14_1:conference},
rather than fine-grained object classification,
has presented conflicting results regarding the impact of camera motion.
Jain et al.~\cite{Jain13_1:conference} showed that features which compensated for camera motion improved performance,
while Kuehne et al.~\cite{Kuehne14_1:conference} showed that the presence of camera motion either had little effect or improved performance.

We manually select 21 classes with videos recorded with and without camera movement,
and examine the performance of the Spatial-DCNN, Temporal-DCNN and the Spatio-Temporal Co-occurrence approach.
The setup of the networks is the same as per Section~\ref{sec:comparative_evaluation}.
The results in Table~\ref{table:camera} show that Spatial-DCNN is adversely affected by camera movement with the accuracy dropping from 57.6\% to 47.8\%. 
This leads to a similar degradation in performance for the Spatio-Temporal Co-occurrence approach: from 61.1\% to 53.7\%.
We attribute the degradation in performance of the spatial networks to the highly challenging conditions,
such as the difference between stationary and flying bird presented in Fig.~\ref{fig:static}.
By contrast, performance of Temporal-DCNN is largely unaffected.

We hypothesise that the Temporal-DCNN is robust to camera movement due to the mean subtraction operation that can reduce the impact of global motion between frames.
To test the above hypothesis we re-trained the Temporal-DCNN without mean subtraction (no zero-norm).
This results in the performance for the Static case reducing from 32.2\% to 28.9\%, while for the Moving case the performance reduced considerably further: from 33.3\% to 23.7\%.
This supports our hypothesis and highlights the importance of the mean subtraction pre-processing stage for temporal features in the presence of camera motion.

\begin{table}[!b]
\small
\setlength{\tabcolsep}{4pt}
\centering
\caption
  {
  Effect of static and moving cameras on performance, using a 21 class subset of the VB100 dataset without bounding box detections.
  Temporal-DCNN (no zero-norm) is trained without applying mean subtraction to the input features.
  }
\label{table:camera}
\begin{tabular}{llc}
\hline\hline\noalign{\smallskip}
{\bf Network} & {\bf Camera Type}  & {\bf Mean Accuracy}\hspace{-1ex} \\
\noalign{\smallskip}
\hline\hline
\noalign{\smallskip}
Spatial-DCNN                   & Static  &  57.6\% \\
Spatial-DCNN                   & Moving  &  47.8\% \\ \hline
Temporal-DCNN (no zero-norm)   & Static  &  28.9\% \\
Temporal-DCNN (no zero-norm)   & Moving  &  23.7\% \\ \hline
Temporal-DCNN                  & Static  &  32.2\% \\
Temporal-DCNN                  & Moving  &  33.3\% \\ \hline
Spatio-Temporal Co-occurrence   & Static  &  \textbf{61.1\%} \\
Spatio-Temporal Co-occurrence   & Moving  &  \textbf{53.7\%} \\
\hline
\end{tabular}
\end{table}

\section{Main Findings}
\label{sec:conclusion}

In this work, we introduced the problem of video-based fine-grained object classification
along with a challenging new dataset and explored methods to exploit the temporal information.
A~systematic comparison of state-of-the-art DCNN based approaches adapted to the task was performed
which highlighted that incorporating temporal information is useful for improving performance and robustness.
We presented a system that encodes local spatial and temporal co-occurrence information, based on the bilinear CNN, that outperforms 3D~ConvNets and the Two-Stream approach.
This system improves the mean classification accuracy from 23.1\% for still image classification to 41.1\%.
Incorporating bounding box information, automatically estimated using the Faster Region CNN, further improves performance to 53.6\%.

In conducting this work we have developed and released the novel video bird dataset VB100 which consists of 1,416 video clips of 100 bird species.
This dataset is the first for video-based fine-grained classification and presents challenges such as how best to combine the spatial and temporal information for classification.
We have also highlighted the importance of normalising the temporal features, using zero-centering, for fine-grained video classification.

Future work will exploit other modalities by incorporating the audio (sound), taxonomy information, and the textual description of the video clips.

\newpage
\section*{Acknowledgement}

\begin{small}
\noindent
The Australian Centre for Robotic Vision is supported by the Australian Research Council via the Centre of Excellence program.
\end{small}

\balance
\bibliographystyle{ieee}
\bibliography{refs}
\end{document}